\documentclass[letterpaper, 10 pt, journal, twoside]{IEEEtran}  % Comment this line out if you need a4paper
\IEEEoverridecommandlockouts                              % This command is only needed if
\usepackage{epsfig} % for postscript graphics files
\usepackage{times} % assumes new font selection scheme installed
\usepackage{hyperref}
\usepackage{mathtools}
\usepackage{xcolor}
\newcommand{\rev}[1]{{\color{black} #1}}
\usepackage{amssymb}  % assumes amsmath package installed
\usepackage{mathrsfs}
\usepackage{comment} % assumes amsmath package installed
\usepackage{algorithm,algorithmic}
\usepackage{amsmath}
\usepackage{empheq}
\usepackage{marginnote}

\title{SLIP Walking over Rough Terrain via H-LIP Stepping and
Backstepping-Barrier Function Inspired Quadratic Program}
    \author{Xiaobin Xiong and Aaron Ames% 
\thanks{Manuscript received Oct. 15, 2020; Revised Jan. 17, 2021; Accepted Feb. 16, 2021. Date of current version: Feb. 18, 2021.}
\thanks{This paper was recommended for publication by Editor Abderrahmane Kheddar upon evaluation of the Associate Editor and Reviewers' comments.
This work was supported by NRI 1924526 and 1923239. \textit{(Corresponding author:
Xiaobin Xiong.)}}
\thanks{The authors are with the California Institute of Technology, Pasadena, CA 91125 USA (email: xxiong@caltech.edu; ames@caltech.edu).}%
        \thanks{This letter has supplementary downloadable material available at http://ieeexplore.ieee.org, provided by the authors.}
\thanks{Digital Object Identifier (DOI): see top of this page.}
 }
\begin{document}
\maketitle

\markboth{IEEE Robotics and Automation Letters. Preprint Version. Accepted Feb., 2021}
{Xiong and Ames: SLIP Walking over Rough Terrain via H-LIP Stepping and Backstepping-Barrier Function} 

\begin{abstract}
We present an \rev{advanced and novel} control method to enable actuated Spring Loaded Inverted Pendulum model to walk over rough and challenging terrains. The high-level philosophy is the decoupling of the controls of the vertical and horizontal states. The vertical state is controlled via Backstepping-Barrier Function (BBF) based quadratic programs: a combination of control Lyapunov \textit{backstepping} and control \textit{barrier} function, both of which provide inequality constraints on the inputs. The horizontal state is stabilized via Hybrid-Linear Inverted Pendulum (H-LIP) based stepping, which has a closed-form formulation. Therefore, the implementation is computationally-efficient. We evaluate our method in simulation, which demonstrates the aSLIP walking over various terrains, including slopes, stairs, and general rough terrains with uncertainties.     
\end{abstract}


\IEEEpeerreviewmaketitle

\section{INTRODUCTION}

\IEEEPARstart{T}{he} Spring Loaded Inverted Pendulum (SLIP) \cite{raibert1986legged, full1999templates} has been a valuable model to study in legged locomotion. Despite its simplicity, it has been used to model and study the dynamics of complex biological locomotion \cite{geyer2006compliant, full1999templates}, and has also inspired control methodologies \cite{raibert1986legged, wuGeyer,xiong2018bipedal, hutter2010slip} and design principles \cite{rezazadeh2018robot, haldane2017repetitive} for high dimensional legged robots. 

The SLIP generates different locomotion behaviors, i.e., hopping/running and walking. For hopping/running, the SLIP has one spring-loaded leg attached to the point mass on the ground phase; it has a linear point mass dynamics in the flight phase. For walking, the (bipedal) SLIP \cite{geyer2006compliant} has two spring-loaded legs, and its dynamics are described by the single support phase (SSP) and double support phase (DSP), based on the number of legs that contact the ground. 

The canonical setting of the SLIP is energy-conservative. The spring has no damping, and there is no energy loss at impact. The control is thus by changing the touch-down angle \cite{geyer2006compliant}, or equivalently the step size \cite{raibert1986legged}. This setting simplifies the control synthesis and analysis; however, it loses certain correspondence to the physical \rev{robots that are designed to resemble the SLIP \cite{GeyerChapter, Ahmadi2006ControlledPD} (namely SLIP-like robots). This is due to the added actuation to compensate for energy dissipation on the real systems}. As a result, either \rev{heuristics-based controllers \cite{raibert1986legged, Ahmadi2006ControlledPD} are directly synthesized on the SLIP-like robots, or actuated versions of the SLIP \cite{ernst2010spring, terry2015control, xiong2018bipedal, green2020planning, rezazadeh2015spring} are proposed for better-approximating the robot dynamics and synthesizing the controllers for their robots}.

\begin{figure}[t]
      \centering
      \includegraphics[width = .9\columnwidth]{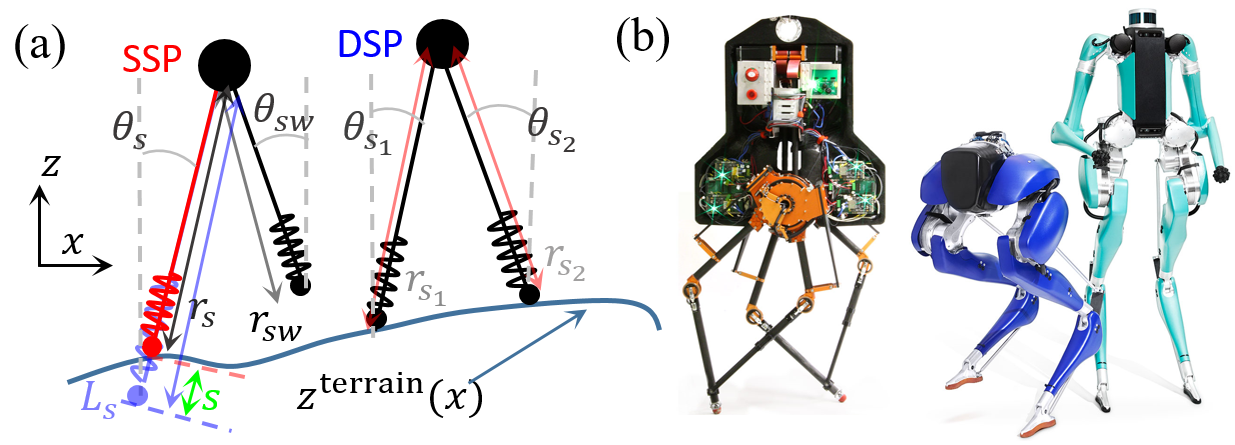}%  inAir.jpg}
      \caption{(a) aSLIP walking on rough terrain, (b) some examples of the aSLIP-like robots: ATRIAS \cite{rezazadeh2015spring}, Cassie \cite{xiong2018bipedal}, Digit (Photo by Dan Saelinger).}
      \label{fig:overview}
\end{figure}

\rev{One common way of actuating the SLIP is via the leg length actuation \cite{liu2016terrain, rezazadeh2015spring, piovan2015reachability,shemer2017flight}, which is in series with the spring. \cite{terry2015control,xiong2018bipedal, green2020planning} further traced back the leg length to its second order dynamics to map the actuated robot dynamics to the SLIP.} We refer to this class of SLIP as the actuated SLIP (aSLIP) \rev{and theirs associated robots as aSLIP-like robots \cite{rezazadeh2015spring, rezazadeh2018robot, dadashzadeh2014template, xiong2018bipedal}: typically with serial-elastic torque-actuated legs (Fig. \ref{fig:overview}).} \rev{The aSLIP is an important model to study, since it has not only been successfully used to synthesize controllers \cite{rezazadeh2015spring,terry2015control,xiong2020jumping,dadashzadeh2014template, xiong2018bipedal} on the aSLIP-like robots, but also been used to provide template dynamics for fully-actuated humanoids \cite{liu2016terrain, xiong2019exo, xiong2020ral}.}

In this paper, we are interested in realizing dynamic walking of the aSLIP on rough terrain. Despite extensive studies on \textit{running} on rough terrain \cite{ wuGeyer,shemer2017flight}, SLIP \textit{walking} on rough terrain has been less studied. The nonlinear dynamics alternating between the SSP and DSP challenge the control, and the influence of the foot-placement on its hybrid dynamics becomes complex. To solve this problem, we decouple the control of the aSLIP walking into two sub-problems: the \textit{continuous} control on the vertical trajectory of the mass via leg internal actuation, and the \textit{discrete} control on the horizontal state of the mass via stepping. 

The continuous control is realized via backstepping \cite{kokotovic} based control Lyapunov function inspired quadratic programs with an integration of control barrier functions \cite{ames2016control}. In short, we name it as Backstepping-Barrier function (BBF). The backstepping solves the hierarchical control from the leg actuation to the leg force and then to the vertical state of the mass. The barrier functions are utilized to keep the leg force remains positive in the SSP and drive one of the leg forces to decrease to 0 in DSP so that it transits to the next SSP at an appropriate timing. The BBF represents affine inequality conditions on the control input, thus quadratic programs (QPs) are formulated for realizing the control. 

The discrete control problem is solved via the stepping based on the Hybrid Linear Inverted Pendulum (H-LIP) \cite{xiong2019IrosStepping, xiong2020ral,xiongInPrep}. The horizontal step-to-step (S2S) dynamics of the aSLIP walking is approximated by that of the H-LIP, where the step size becomes the input. we extend the results of the H-LIP stepping on flat terrain in \cite{xiong2020ral} to the case of walking on rough terrains. The stepping keeps the discrete horizontal state of the aSLIP close to that of the H-LIP, the error between which converges an error invariant set. Desired walking behaviors are thus approximately realized on the aSLIP.

\rev{Compared to the approaches in \cite{dai2012optimizing, manchester2011stable, nguyen2018dynamic} for walking on rough terrains, the proposed approach is highly efficient in computation and easy to implement.} The continuous controls via BBF-QPs are convex optimizations. The discrete control via H-LIP stepping is in closed-form. \rev{Additionally, compared to \cite{manchester2011stable,iida2010minimalistic, liu2016terrain}, our controller tolerates significantly larger height variations of the terrain.} Moreover, the terrain does not need to be exactly known, thus robustness is promoted. 

We evaluate the proposed control scheme on walking on rigid terrains with various shapes and uncertainties. The aSLIP can achieve desired walking on all tested terrains successfully. We envision this approach to be extended for aSLIP-like robots and to provide template walking dynamics for fully-actuated humanoids to walk on rough terrains. 

\section{The aSLIP Model of Walking}
In this section, we present the dynamics of the walking of the actuated Spring Loaded Inverted Pendulum (aSLIP) and then present the control methodology at the high-level.

\subsection{Dynamics of Walking}
The aSLIP (Fig. \ref{fig:overview} (a)), similar to the canonical SLIP \cite{geyer2006compliant}, contains a point mass attached on two prismatic springy legs. The walking alternates between two domains: the Single Support Phase (SSP) and the Double Support Phase (DSP). The point mass moves under the leg forces and the gravitation. The point mass dynamics is: \begin{equation}
 m \ddot{\mathbf{P}} = \textstyle \sum \mathbf{F} + m\mathbf{g},
\end{equation} 
where $m$ is the mass, $\mathbf{P} = [x, z]^T$ is the position vector of the mass, $\mathbf{F}$ represents the leg forces on the stance legs, and $\mathbf{g}$ is the gravitation vector. It transits from the DSP to the SSP when one of the legs is about to lift off (the leg force is crossing zero). The walking transits into the DSP when the swing foot strikes the ground. One important control input of aSLIP and the only input of the canonical SLIP are the swing leg angle in the SSP, which can be directly set since the swing leg is assumed to be mass-less.

The leg length actuation on the aSLIP provides additional control inputs. The actuation can extend and retract the uncompressed leg length $L$. Let $r$ be the compressed leg length and $s$ be the spring deformation, thus $L = s + r$. The actuation of the leg length is $\ddot{L} = \tau$, where $\tau$ is the input.

We also add damping on the spring for energy dissipation. Then, the leg force is $\rev{F_s} = \text{K} s + \text{D} \dot{s}$, where $\text{K}$ and $\text{D}$ are the spring stiffness and damping{\footnote{\rev{  $\text{K}$ and $\text{D}$ of the leg spring on the physical robot can be nonlinear \cite{xiong2018bipedal} of $L$ for best approximation or constant for simplifications.}}}, respectively. As a result, its system dynamics can be written in polar coordinates, e.g., the stance leg dynamics in SSP are:
\begin{align}
 \ddot{r}_s &= \textstyle \tfrac{F_s}{m} - g \text{cos}(\theta_s) + r_s \dot{\theta}_s^2  \nonumber \\
\ddot{\theta}_s &=\textstyle \tfrac{1}{r_s} ( -2\dot{\theta}_s \dot{r}_s + g \text{sin}(\theta_s) ) \nonumber \\
 \ddot{s}_s&= \textstyle \tau_s - \ddot{r}_s \nonumber 
\end{align}
where $\theta$ is the leg angle, and the subscript $_s$ denotes the stance leg. In the latter, we use $_{sw}$ to denote the swing leg in the SSP. In DSP, the subscript $_{s_1}$ and $_{s_2}$ denote the two stance legs and $_{s_1}$ is the leg that will lift off from the ground. Using the subscripts, the vertical acceleration in DSP is:
\begin{align}
\label{eq:verticaldynamics}
 \textstyle    \ddot{z} = \frac{F_{s_1}}{m} \text{cos}(\theta_{s_1}) + \frac{F_{s_2}}{m} \text{cos}(\theta_{s_2}) - g.
\end{align}

\subsection{Control Scheme for Walking on General Rigid Terrain}

We consider that the walking requires \rev{\textit{three specifications} as shown in Fig. \ref{fig:controloverview} (a)}: it keeps a vertical distance from the ground, the swing foot periodically lifts off and strikes the ground to switch support legs, and the swing foot steps to certain locations to produce a desired horizontal behavior. The corresponding controls are briefly explained as follows.

\subsubsection{Vertical Mass Control via BBF-QP} 
The vertical state $z$ is expected to follow a desired trajectory $z^d$ which has an approximately constant distance from the ground (see Fig. \ref{fig:controloverview}). The leg forces are expected to be positive and $F_{s_1}$ has to cross 0 at the end of the DSP. The vertical tracking and the leg force conditions are solved via the Backstepping-Barrier function based quadratic programs (BBF-QPs), which will be explained in section \ref{sec:backstepping} and \ref{sec:barrier}.

\subsubsection{Vertical Swing Foot Control}
 The vertical position of the swing foot is controlled to lift off, avoid scuffing, and strike on the ground to finish the SSP at appropriate timing. The desired vertical swing foot position is constructed as:
\begin{align}
\label{eq:swingZpos}
    z_{sw}^{d} = z^{\text{time}}_{sw}(t) + z^{\text{terrain}}(x_{sw}),
\end{align}
where $z^{\text{time}}_{sw}(t)$ is the time-dependent component and $z^{\text{terrain}}(x_{sw})$ is the terrain profile. $z^{\text{time}}_{sw}(t)$ is constructed so that the swing foot lifts off from the ground, reaches to a maximum height $z^{\text{max}}_{sw}$ and then strikes the ground at $t = T_\text{S}$, where $T_\text{S}$ is the duration of the SSP. An example of the time-dependent component is:
$z^{\text{time}}_{sw}(t) = z^{\text{max}}_{sw} \text{cos}(\frac{t}{T_\text{S}} \pi - \frac{1}{2}\pi)$. The spring compression is assumed to go to zero on the swing leg, thus we select $z_{sw} = z - L_{sw}\text{cos}(\theta_{sw})$ and apply a feedback linearizing controller to drive $z_{sw} \rightarrow z_{sw}^{d}$.

Note that the terrain profile may not be exactly known in practice, for which we assume an estimated version is available. The uncertainty on the terrain height creates an uncertainty on the duration of the SSP: early strike causes a shorter duration and late strike produces a longer duration. 

\subsubsection{Horizontal Mass Control via H-LIP Stepping}
\begin{figure}[t]
      \centering
      \includegraphics[width = 1\columnwidth]{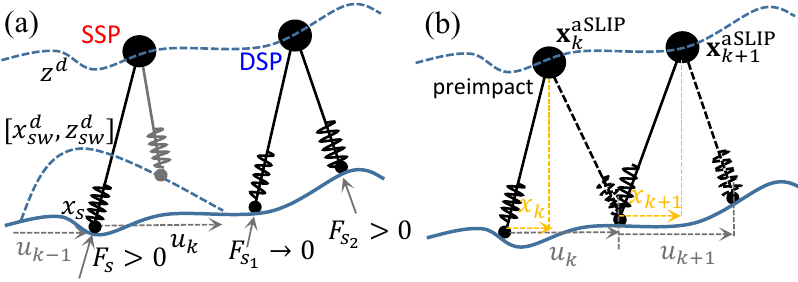}
      \caption{(a) The control specifications and (b) the step-to-step dynamics. }
      \label{fig:controloverview}
\end{figure}

The horizontal state $\mathbf{x} = [x, \dot{x}]^T$ should be controlled for walking. Due to the point-foot underactuation, the horizontal state cannot be continuously controlled. We stabilize the state at the \textit{pre-impact event} based on the discrete step-to-step (S2S) dynamics \cite{xiong2020ral} of the walking (see Fig. \ref{fig:controloverview} (b)). The step size $u$ is considered as the input to the S2S of the discrete horizontal state. The S2S dynamics is approximated by the S2S dynamics of the H-LIP. The H-LIP based stepping is applied to provide the desired step size on the aSLIP, which is explained in section \ref{sec:HLIP}. Since the swing angle can be directly set, we simply construct a smooth trajectory of the horizontal swing foot position to transit from previous step location to the desired location. Then the horizontal swing foot position $x^d_{sw}$ is: 
\begin{align}
\label{eq:swingXpos}
    x^d_{sw} = x_{s} + c(t) u_k - (1-c(t)) u_{k-1},
\end{align}
where $x_{s}$ is the horizontal position of the stance foot, $c(t)$ is a smooth time-based curve to transit from 0 to 1 within the SSP duration $T_\text{S}$, and $u_{k-1}$ is the previous step size.

\section{Backstepping-Control Lyapunov Function for
Vertical Stabilization}
\label{sec:backstepping}
In this section, we describe one of the main components of this paper for controlling the vertical behavior of the walking of the aSLIP. We present the dynamics structure of the vertical state in each domain. Then we show the canonical Lyapunov backstepping that \rev{\textit{guarantees} to stabilize} this class of control problem \rev{without heuristic gain-tuning}. Last, we formulate a control Lyapunov backstepping, which yields an inequality condition for the control \rev{and opens opportunities to include extra constraints on the input in an optimization-based controller}. 

\subsection{Strict-feedback Form of the Vertical State}
The objective is to drive the vertical position of the mass to follow a desired trajectory. We define the output as:
\begin{equation}
     \eta := \begin{bmatrix}
        z - z^{d}(t),\dot{z} - \dot{z}^d(t) 
\end{bmatrix}^T,
\end{equation}
where $[z^{d}(t), \dot{z}^d(t)]$ is the desired trajectory to follow. Differentiating the output yields the output dynamics:
\begin{align}
\label{eq:eta_dynamics}
\dot{\eta} = f_\eta  + g_\eta F_z^{\text{SSP/DSP}}, 
\end{align}
where $ f_\eta:=  \begin{bmatrix} \dot{z} - \dot{z}^d, - g - \ddot{z}^d 
\end{bmatrix}^T,~ g_\eta := \begin{bmatrix} 0, \textstyle\frac{1}{m}
\end{bmatrix}^T$. 
$F_z^{\text{SSP/DSP}}$ is the net vertical force in each domain: 
\begin{align}
  F^{\text{SSP}}_z = F_{s}\text{cos}(\theta_{s}),~
 F^{\text{DSP}}_z & = F_{s_1} \text{cos}(\theta_{s_1})  + F_{s_2} \text{cos}(\theta_{s_2}). \nonumber
\end{align}
We view $F_z^{\text{SSP/DSP}}$ as the fictitious input to this system in Eq. \eqref{eq:eta_dynamics}. Differentiating $F_z^{\text{SSP/DSP}}$ yields the affine dynamics with the actual actuation of the leg length as the input. The dynamics are different in the SSP and DSP:
\rev{\begin{align}
\label{eq:F_dynamics}
\dot{F}_z^{\text{SSP}} = f^{\text{SSP}}_z + g_z \tau_z^{\text{SSP}},~\dot{F}_z^{\text{DSP}} = f^{\text{DSP}}_z + g_z \tau_z^{\text{DSP}},
\end{align}}
where $g_z =  \text{D} \neq 0$, and, 
\begin{equation}
       \label{eq:tau_z_SSPDSP}
   \tau^{\text{SSP}}_z := \text{cos}(\theta_s) \tau_s,~
    \tau^{\text{DSP}}_{z}:=  \text{cos}(\theta_{s_1})  \tau_{s_1} + \text{cos}(\theta_{s_2}) \tau_{s_2}.
\end{equation}
The expressions of $f^{\text{SSP}}_z$ and $f^{\text{DSP}}_z$ are omitted. $\tau_z^{\text{SSP/DSP}}$ represents the vertical component of the leg length actuation. The leg angles $\theta \in (-\frac{\pi}{2}, \frac{\pi}{2})$; thus, given a desired $\tau_z^{\text{SSP/DSP}}$, there always exists leg length actuation for realization. Note that the desired $\tau_z^{\text{DSP}}$ is not uniquely realized by $\tau_{s_1}$ and $\tau_{s_2}$.

As a result, for walking in both domains, we can synthesize $\tau_z \in \mathbb{R}$ to stabilize the vertical trajectory of the mass based on the dynamics in the \textit{strict-feedback form} \cite{khalil1996noninear}:
\begin{empheq}[box=\fbox]{align}
\label{eq:strickfb1}
    \dot{\eta} &= f_\eta + g_\eta F_z,  \\
\label{eq:strickfb2}
\dot{F}_z &= f_z + g_z \tau_z.
\end{empheq}

\subsection{Lyapunov Backstepping}
Lyapunov Backstepping \cite{kokotovic} can be applied to stabilize the dynamics in Eq. \eqref{eq:strickfb1} and \eqref{eq:strickfb2}. For Eq. \eqref{eq:strickfb1} with $F_z$ being the input, a feedback linearizing controller can be synthesized: 
\begin{align}
    \bar{F}_z =\textstyle \frac{1}{{g_\eta}_2}(-{f_\eta}_{2} + K_{\text{IO}} \eta),
\end{align}
where $K_{\text{IO}} = [K_p, K_d]$ is the linear feedback gain and the subscript $_2$ indicates the second element of the vector. This yields the linear closed-loop dynamics: 
\begin{align}
    \dot{\eta} =f_{\eta} + g_{\eta} \bar{F}_z =  \begin{bmatrix} 0 & 1 \\
    K_p & K_d \end{bmatrix} \eta := A_{cl} \eta.
\end{align}
$K_{\text{IO}}$ is chosen with $K_p <0, K_d < 0$ so that $A_{cl}$ is stable (with negative eigenvalues). On the closed-loop dynamics, a Lyapunov function can be found: $V_{\eta} = \eta^T P \eta$, with $P>0$ (being positive definite). $P$ satisfies the continuous-time Lyapunov function $P A_{cl} + A^T_{cl}P = -Q$, where $Q>0$ is selected. It is easy to verify that: $\dot{V}_{\eta} = - \eta^T Q \eta \leq -\lambda_{\text{min}}(Q)||\eta||^2$, where $\lambda_{\text{min}}(Q)$ is the smallest eigenvalue of the matrix $Q$.

To synthesize the actual control from $\tau_z$ to stabilize $\eta \rightarrow 0$, we define $ F_{\delta} := F_z - \bar{F}_z$ and the Lyapunov function be:
\begin{align}
    V(\eta, F_z) = V_{\eta} + \textstyle\frac{1}{2} F_{\delta}^2.
\end{align}
Differentiating this yields, 
\begin{align}
   \dot{V}(\eta, F_z) &=  \textstyle \frac{\partial V_{\eta}}{\partial\eta}(f_{\eta} +g_{\eta} F_z) + F_{\delta} \dot{F_\delta}\nonumber \\
     &= \textstyle  \frac{\partial V_{\eta}}{\partial\eta}(f_{\eta} +g_{\eta} \bar{F}_z) + \frac{\partial V_{\eta}}{\partial\eta}g_{\eta} F_{\delta} + F_{\delta} \dot{F_\delta}\nonumber\\
     &=  \textstyle \dot{V}_\eta + \frac{\partial V_{\eta}}{\partial\eta}g_{\eta} F_{\delta} + F_{\delta} \dot{F_\delta} \label{eq:VdotAffine}\\
     &\textstyle  \leq -\lambda_{\text{min}}(Q) ||\eta||^2 + \frac{\partial V_{\eta}}{\partial\eta}g_{\eta} F_{\delta} + F_{\delta} \dot{F_\delta} .
\end{align}
 If we choose 
\begin{align}
\label{eq:closedControlF_delta}
   \textstyle  \dot{F_\delta} = - \frac{\partial V_{\eta}}{\partial\eta}g_{\eta} - k F_{\delta},
\end{align}
with $k>0$, then,
\begin{align}
     \dot{V}(\eta, F_z) &\leq -\lambda_{\text{min}}(Q) ||\eta||^2  - k F_\delta^2 \nonumber \\ 
    & \leq - \text{min}(\lambda_{\text{min}}(Q), k) ||[\eta^T ,F_\delta]^T||^2.
\end{align}
By Lyapunov's method, the system with $(\eta, F_\delta)$ as states is exponentially stable to the origin $(\eta, F_\delta) = (\mathbf{0}, 0)$. This provides a closed-form \textit{Backstepping controller} on $\tau_z$ from Eq. \eqref{eq:closedControlF_delta}. Since
$ \dot{F_\delta} = f_z + g_z \tau_z - \dot{\bar{F}}_z$, the controller is 
\begin{align}
\label{eq:backstepping-closedform}
      \tau_{z} = \frac{1}{g_z}(- \frac{\partial V_{\eta}}{\partial\eta}g_{\eta} - k F_{\delta} +  \dot{\bar{F}}_z - f_z).
\end{align}

\subsection{Backstepping-CLF}
The closed-form controller in Eq. \eqref{eq:backstepping-closedform} appears to be able to stabilize $\eta \rightarrow 0$. However, the resultant leg force $F_z$ can be negative, which is not valid for walking. Here we develop the control Lyapunov function (CLF) of the backstepping to provide an \textit{inequality condition} on the input $\tau_z$ to stabilize $\eta$. The inequality motivates an optimization-based controller, in which the condition of non-negative leg forcing can be enforced via additional constraints, e.g., the control barrier function (CBF) in the next section.  

Note that $\dot{V}$ is an affine function w.r.t. the input $\tau_z$: 
\begin{align}
    \dot{V}(\eta, F_z) &=  \dot{V}_\eta + \frac{\partial V_{\eta}}{\partial\eta}g_{\eta} F_{\delta} + F_{\delta}(f_z + g_z \tau_z - \dot{\bar{F}}_z) \nonumber \\
    & =\dot{V}_\eta + \frac{\partial V_{\eta}}{\partial\eta}g_{\eta} F_{\delta}  + F_{\delta}(f_z - \dot{\bar{F}}_z) + F_{\delta} g_z \tau_z. \nonumber
\end{align}
The exponential stability can be established via enforcing:
\begin{align}
    \dot{V}(\eta, F_z) \leq -\gamma V(\eta, F_z),
\end{align}
with $\gamma > 0$. This yields a \textit{Backstepping-CLF inequality}:
\begin{empheq}[box=\fbox]{align}
\label{eq:backstepping-clf}
A^{\text{Backstepping}}_{\text{CLF}} \tau_z \leq b^{\text{Backstepping}}_{\text{CLF}} 
\end{empheq}
where $A^{\text{Backstepping}}_{\text{CLF}}:=  F_{\delta} g_z, b^{\text{Backstepping}}_{\text{CLF}}:=  -\dot{V}_\eta - \frac{\partial V_{\eta}}{\partial\eta}g_{\eta} F_{\delta}  - F_{\delta}(f_z - \dot{\bar{F}}_z) - \gamma V. $
When $F_\delta \neq0$, Eq. \eqref{eq:backstepping-clf} is a constraint on $\tau_z$. When $F_\delta = 0$, the inequality becomes $\dot{V}_\eta \leq - \gamma V =  - \gamma V_\eta$, which is automatically satisfied as long as $0 \leq \gamma \leq \frac{\lambda_\text{min}(Q)}{\lambda_\text{max}(P)}$.
As a result, as long as $\tau_z$ satisfies the backstepping-CLF inequality, $\eta$ exponentially converges to 0. Note that this inequality is an affine condition on $\tau_{s}$ in SSP or $\tau_{s1}$ and $\tau_{s2}$ in DSP as indicated by Eq. \eqref{eq:tau_z_SSPDSP}. Thus, in the next section, we will formulate quadratic program (QP) based controllers that include the inequality in Eq. \eqref{eq:backstepping-clf} with the incorporation of the control barrier functions. 

%%%%%%%%%%%%%%%%%%

\section{Control Barrier Functions for Walking}
\label{sec:barrier}
In the application of walking, the leg forces should be positive during contact. Moreover, in DSP, one leg force should gracefully cross 0 to initiate lift-off. These conditions can be \rev{described via sets and thus be} enforced via control barrier function (CBF) \rev{with an inequality condition which \textit{guarantees} set invariance on the dynamics}. We start by introducing the CBF, show the application for the walking of the aSLIP, and finally integrate it with the Backstepping-CLF to formulate the final backstepping-barrier function based quadratic program (BBF-QP) controllers for walking.

\subsection{Control Barrier Functions}
The control barrier function \cite{ames2016control} describes a condition for the control input that guarantees set invariance. We consider a super level set $\mathcal{C}$ of a continuously differentiable scalar function $h: \mathbb{R}^n \rightarrow \mathbb{R}$. By definition: 
$    \mathcal{C}    = \{x \in \mathbb{R}^n| h(x) \geq 0 \}.
$ Here we use $x$ for a general state representation, instead of the horizontal position of the aSLIP. $h$ is a \textit{control barrier function} for the affine control system $\dot{x} = f(x) + g(x) u$ if:
\begin{align}
\label{eq:cbf_def}
    \underset{u\in U}{\text{sup}}[ \mathcal{L}_fh(x) + \mathcal{L}_gh(x) u + \alpha(h(x)) ] \geq 0,
\end{align}
where $\mathcal{L}_{f}h(x) = \frac{\partial h}{ \partial x} f(x)$ and $\mathcal{L}_{g}h(x) = \frac{\partial h}{ \partial x} g(x)$ are the Lie derivatives; $\mathcal{L}_{f}h(x)  + \mathcal{L}_{g}h(x)  = \dot{h}(x)$. $U$ is the set where the input $u$ is in, and $\alpha(\cdot)$ is an extended class $\mathcal{K}_{\infty}$ function\footnote{$ \alpha:\mathbb{R} \rightarrow \mathbb{R}$, $\alpha(0) = 0$ and $\alpha$ is strictly monotonically increasing.}. This condition indicates that there exists an input to stabilize the set $ \mathcal{C}$, i.e., making sure $h(x) \geq 0$. If the state is in $\mathcal{C}$, it will stay in the set forever if the \textit{CBF inequality} is satisfied: 
\begin{empheq}[box=\fbox]{align}
\label{eq:CBF_condi}
  \mathcal{L}_fh(x) + \mathcal{L}_gh(x) u \geq -\alpha(h(x)).
\end{empheq}
This makes sure that the lower bound of the derivative $\dot{h}$ is increasing with the decrease of $h$. It can be proven that the set $\mathcal{C}$ is exponentially stable under this condition \cite{ames2016control}.

\subsection{Application to aSLIP Walking}
Eq. \eqref{eq:CBF_condi} represents an inequality constraint on the input to make sure that $h \geq 0$, for which $h$ is defined differently for the walking in the SSP and the DSP. 

\noindent{{\textbf{SSP:}}}
The stance leg force $F_s$ should be non-negative, so is its vertical component $F^{\text{SSP}}_z$. Thus, we let:
\begin{equation} 
   h_s = F^{\text{SSP}}_z, \quad \dot{h}_s \geq - \alpha (h_s),
\end{equation}
which provides an inequality on the input $\tau_s$:
\begin{align}
    A^{s}_{\text{CBF}} \tau_s \leq b^{s}_{\text{CBF}},
\end{align}
where $\rev{A^{s}_{\text{CBF}}:= -g_z \text{cos}(\theta_s)},~b^{s}_{\text{CBF}}:= f^{\text{SSP}}_z + \alpha (h_s)$. We simply select $\alpha(\cdot)$ to be a linear function. This inequality naturally fits with the Backstepping-CLF inequality in Eq. \eqref{eq:backstepping-clf} to formulate a backstepping-barrier function based quadratic program (BBF-QP) controller:
\begin{empheq}[box=\fbox]{align}
\label{eq:BBF_QP_SSP}
 (\tau_s, \delta) &= \underset{(\tau_s,\delta) \in \mathbb{R}^2}{\text{argmin}} \ \tau_{s}^2  + \delta^2\\
     \text{s.t.}& \quad A^{\text{Backstepping}}_{\text{CLF}} \tau^{\text{SSP}}_z \leq b^{\text{Backstepping}}_{\text{CLF}} + \delta \nonumber \\
       &  \quad A^s_{\text{CBF}} \tau_s \leq b^s_{\text{CBF}} \nonumber
\end{empheq}
where $\delta$ is a relaxation variable to avoid infeasibility. In case when the CBF constraint violates the Backstepping-CLF constraint, the Backstepping-CLF constraint is relaxed and the CBF constraint is still enforced. 

\noindent{{\textbf{DSP:}}}
Both leg forces should remain non-negative. The leg force on $s_2$ should remain non-negative through out the DSP. Thus, we let $h_{s_2} = F_{s_2}$, and the CBF inequality becomes
\begin{align}
\label{eq:CBF_DSP_s2_condi}
    A^{s_2}_{\text{CBF}} \tau_{s_2} \leq b^{s_2}_{\text{CBF}}.
\end{align}

The leg force on $s_1$ should gradually decrease and reach to zero to trigger the transition into the next SSP. Let $F^0_{s_1}$ be the leg force on $s_1$ in the beginning of the DSP. A desired leg force trajectory can be designed: $ F^d_{s_1}(t) = F^0_{s_1} (1-\frac{t}{T_\text{D}} )$, where $T_\text{D}$ is the duration of the DSP. One may consider to design a feedback controller to drive $F_{s_1} \rightarrow  F^d_{s_1}$. However, this creates a high restriction on $\tau_{s_1}$ and can lead to conflicts between the Backstepping-CLF inequality in Eq. \eqref{eq:backstepping-clf} and the CBF inequality in Eq. \eqref{eq:CBF_DSP_s2_condi}. 

\begin{figure}[t]
      \centering
      \includegraphics[width = 0.9 \columnwidth]{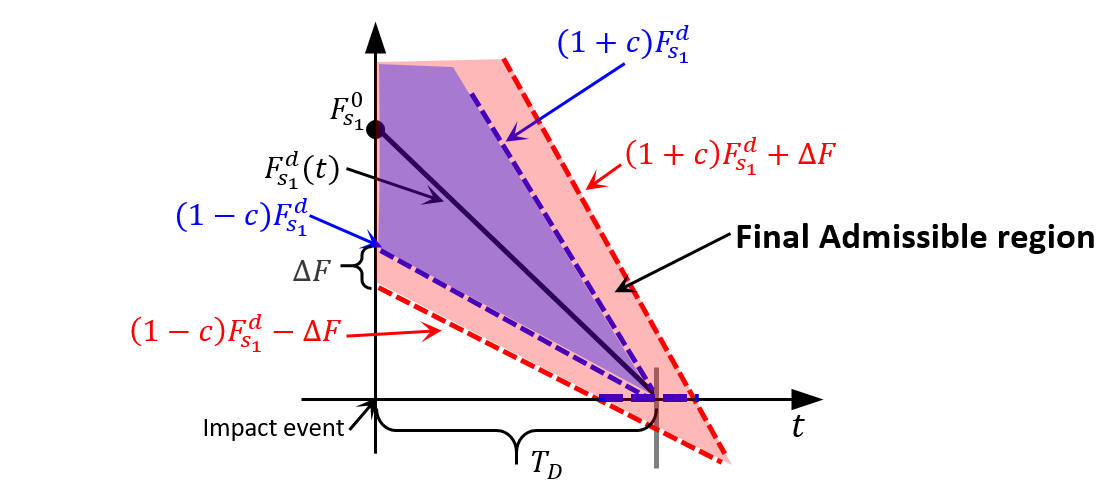}
      \caption{\rev{Contact force condition for lift-off in DSP (the red region represents the admissible region for the leg force).}}
      \label{fig:DSPforceCondi}
\end{figure}

To decrease $F_{s_1}$ in a relaxed fashion, we create the inequality condition: $(1-c)F^d_{s_1} \leq    F_{s_1} \leq (1+c)F^d_{s_1}$, where $c \in(0, 1)$ is a relaxation coefficient. As shown in Fig. \ref{fig:DSPforceCondi}, this generates an admissible force region (indicated by the blue region), which decreases as the desired force $F^d_{s_1}$ decreases with time. This two inequalities can be converted into a single inequality: 
$h_{s_1} = (cF^d_{s_1})^2 - (F_{s_1} - F^d_{s_1})^2 \geq 0$. Note that this barrier function is ill-defined as $F^d_{s_1}$ approaches to 0 (the set $\mathcal{C}$ becomes trivial). Thus, we increase the relaxation by adding a positive value $\Delta F$ in the inequality: 
\begin{align}
(1-c)F^d_{s_1} - \Delta F \leq    F_{s_1} \leq (1+c)F^d_{s_1} + \Delta F,
\end{align}
which generates the red admissible region. By defining:
\begin{align}
    h_{s_1} = (cF^d_{s_1}  + \Delta F)^2 - (F_{s_1} - F^d_{s_1} )^2 \geq 0 ,
\end{align}
the set $\mathcal{C}$ is always non-trivial before lift-off. Thus we have another CBF inequality: $ A^{s_1}_{\text{CBF}} \tau_{s_1} \leq b^{s_1}_{\text{CBF}}$.

Similarly, the two CBF inequalities are incorporated with the Backstepping-CLF inequality to formulate the final BBF-QP controller for the walking in DSP:
\begin{empheq}[box=\fbox]{align}
\label{eq:BBF_QP_DSP}
    (\tau_{s1},\tau_{s2}, \delta)& = \underset{(\tau_{s1},\tau_{s2}, \delta) \in \mathbb{R}^3}{\text{argmin}} \ \tau_{s1}^2  + \tau_{s2}^2 + \delta^2 \\
    \text{s.t.} & \quad A^{\text{Backstepping}}_{\text{CLF}} \tau^{\text{DSP}}_z
    \leq b^{\text{Backstepping}}_{\text{CLF}} + \delta \nonumber \\
      & \quad    A^{s_1}_{\text{CBF}} \tau_{s_1} \leq b^{s_1}_{\text{CBF}},     A^{s_2}_{\text{CBF}} \tau_{s_2} \leq b^{s_2}_{\text{CBF}}\nonumber.
\end{empheq}

The BBF-QPs are designed to stabilize the vertical position of the mass to the desired trajectory and simultaneously satisfy the conditions on the leg forces during walking.

%%%%%%%%%%%%%%%%%%%%%%%%%%%%%%%%%%%%%%%%%%%%%%%%%%%%%%%%%%%%%%%%%%%%%%
%%%%%%%%%%%%%%%%%%%%%%%%%%%%%%%%%%%%%%%%%%%%%%%%%%%%%%%%%%%%%%%%%%%%%%
%%%%%%%%%%%%%%%%%%%%%%%%%%%%%%%%%%%%%%%%%%%%%%%%%%%%%%%%%%%%%%%%%%%%%%
%%%%%%%%%%%%%%%%%%%%%%%%%%%%%%%%%%%%%%%%%%%%%%%%%%%%%%%%%%%%%%%%%%%%%%

\section{H-LIP Stepping For Horizontal Stabilization}
\label{sec:HLIP}
We now describe the horizontal stabilization via stepping, which is based on the \textit{step-to-step} (S2S) dynamics approximation via the \textit{Hybrid Linear Inverted Pendulum} (H-LIP) \cite{xiong2019IrosStepping, xiong2020ral}. The H-LIP based stepping has been proposed in \cite{xiong2019IrosStepping,xiong2020ral, xiongInPrep} for walking on flat terrain. Here, we generalize the stepping for walking on non-flat terrain.

\subsection{The H-LIP Model and its Step-to-step Dynamics}
\begin{figure}[b]
      \centering
      \includegraphics[width = 0.85 \columnwidth]{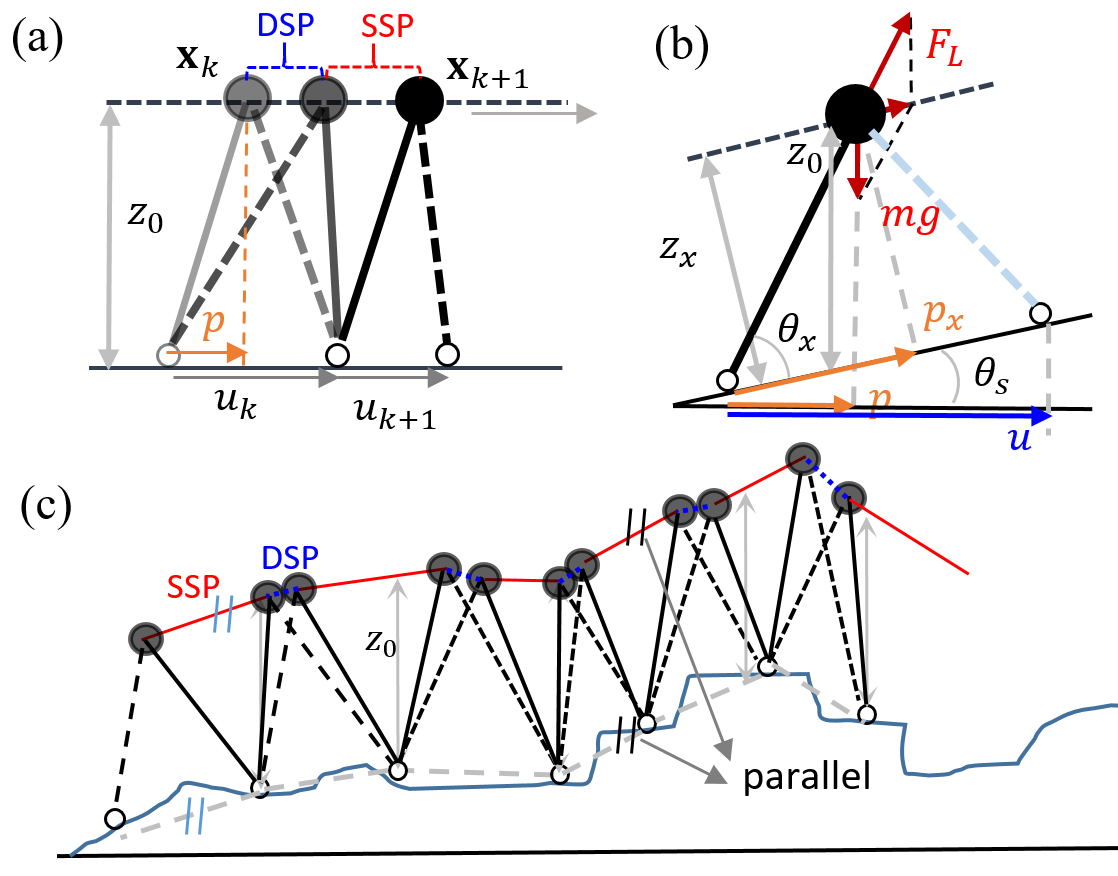}%  inAir.jpg}
      \caption{Illustration of the walking on the H-LIP model on a flat terrain (a) and on a slope (b). (c) H-LIP walking on rough terrain, where the dashed lines indicate the leg that is about to lift off or strike the ground.}
      \label{fig:HLIP}
\end{figure}

The H-LIP \cite{xiong2019IrosStepping} is a formal \rev{adaptation} of the Linear Inverted Pendulum Model in \cite{kajita2002}. It is assumed to walking with a constant height of the mass, telescopic legs, and point feet (see Fig. \ref{fig:HLIP} (a)). The walking is composed of two alternating phases, i.e., the Single Support Phase (SSP) and the Double Support Phase (DSP). The velocity in DSP is assumed to be constant \cite{feng2016robust, kajita2001}. The dynamics are:
\begin{align}
\label{eq:HLIP_ssp}
\text{SSP:} \quad \ddot{p} = \lambda^2 p,  \quad
\text{DSP:}\quad  \ddot{p} = 0,
\end{align}
where $p$ is the horizontal position from the support foot to the mass, $\lambda = \textstyle \sqrt{g/z_0}$, and $z_0$ is the constant height of the mass. The transitions between the phases are smooth. The walking is controlled by changing the step size $u$. \cite{xiong2020ral} proposed a \textit{Step-to-Step (S2S)} dynamics formulation, which treats the step size as the input to discrete dynamical system at the step level. Consider the \rev{\textbf{\textit{pre-impact state}}} (the state before the transition from SSP to DSP) $\mathbf{x}_k = [p_k, \dot{p}_k]^T$ at the step indexed by $k$ with the step size being $u_k$. The \textit{pre-impact state} at the next step follows the S2S dynamics: 
\begin{equation}
\label{eq:S2S}
\mathbf{x}_{k+1} =A  \mathbf{x}_{k} + B u_k,
\end{equation}
where \rev{
\scalebox{0.7}{$
A = \begin{bmatrix}
\textstyle \mathrm{cosh}(T_{\textrm{S}} \lambda) & T_{\textrm{D}} \mathrm{cosh}(T_{\textrm{S}} \lambda) + \frac{1}{\lambda}\mathrm{sinh}(T_{\textrm{S}}\lambda)\\ 
\lambda \mathrm{sinh}(T_{\textrm{S}} \lambda) &  \mathrm{cosh}(T_{\textrm{S}} \lambda) + T_{\textrm{D}} \mathrm{sinh}(T_{\textrm{S}} \lambda)
\end{bmatrix}, B = \textstyle \begin{bmatrix}
 - \mathrm{cosh}(T_{\textrm{S}} \lambda)\\ 
- \lambda \mathrm{sinh}( T_{\textrm{S}} \lambda)
\end{bmatrix}.$}} \\
As a result, we generate desired walking behaviors by controlling the discrete pre-impact state via changing the step size. The periodic walking behaviors can be described in closed-form \cite{xiong2019IrosStepping, xiongInPrep}, and the non-periodic walking can be found via optimization \cite{xiong2020ral}. 

\subsection{Generalization on Walking on Rough Terrain}
We extend the H-LIP walking to rough terrains. We first consider the H-LIP walking on a slope, as shown in Fig. \ref{fig:HLIP} (b). We assume that the vertical distance between the mass and the slope, i.e. $z_0$, remains constant. $\theta_s$ is the degree of the slope. $\theta_x$ is the angle between the leg and the slope. The step size $u$ is defined as the horizontal distance between the feet.

\noindent{\underline{\textit{SSP dynamics:}}} In SSP, the H-LIP is the LIP in \cite{kajita2001} with zero ankle torque. The results of Eq. (12) in \cite{kajita2001} can be applied and yield the identical dynamics in Eq. \eqref{eq:HLIP_ssp}. It can also be derived from the \rev{Newton}-Euler equations: 
\begin{align}
    m \ddot{p}_x  = F_L \text{cos}(\theta_x) - m g \text{sin}(\theta_s),
    F_L \text{sin}(\theta_x)  = m g \text{cos}(\theta_s). \nonumber
\end{align}
where $F_L$ is the leg force. Solving for $\ddot{p}_x$ yields $\ddot{p}_x = \lambda^2 p_x - g \text{sin} (\theta)$.
As $p = p_x \text{cos}(\theta) - z_x \text{sin} (\theta)$, $ \ddot{p} =  \ddot{p}_x \text{cos}(\theta) =  \lambda^2 p$.

\noindent{\underline{\textit{DSP dynamics:}}} In DSP, two leg forces act on the point mass to yield the net acceleration of the mass pointing to the direction which is parallel to the slope. The Newton-Euler equations cannot yield deterministic net accelerations. In other words, the net acceleration is controllable from the leg forces. Thus we assume that the acceleration is zero. 

\noindent{\underline{\textit{S2S dynamics:}}} As a result, by describing the state and input in the inertial frame, the dynamics of the H-LIP walking on the slope is identical to the dynamics on flat ground, so is the step-to-step dynamics and the resulting stepping controller. 

\noindent{\underline{\textit{On rough terrain:}}} The walking model of the H-LIP on general non-flat terrains is based on the walking on slopes. Given a sequence of steps of the H-LIP walking on rough terrain, the walking is equivalent to walking on piecewise continuous slopes, as illustrated in Fig. \ref{fig:HLIP} (c). Since the slope changes with each step, the assumption has to be made to enable the change of the slope. For instance, in \cite{kajita1991study}, an impulse of the leg force is assumed to change the slope rate when the leg is strictly vertical in the SSP. Here we assume the slope changes in DSP, where the two leg forces can simultaneously create zero horizontal acceleration and change the vertical trajectory. Therefore, the dynamics in both domains remain the same, so does the S2S dynamics. \rev{The assumptions on the H-LIP walking are designed to have a linear S2S dynamics to approximate and control the actual walking of the aSLIP model, which is explained as follows.}

\subsection{Stepping Control based on the S2S Approximation}
After we verify that the S2S dynamics of the H-LIP on rough terrain can still be a linear dynamics, we use it as an approximation to the actual S2S dynamics of the aSLIP walking. Here we briefly explain the stepping controller based on the S2S approximation. More details are in \cite{xiong2020ral}.

Consider the S2S of the horizontal aSLIP state as: 
\begin{align}
    \label{eq:SLIP_S2S}
\mathbf{x}^{\text{aSLIP}}_{k+1} = \mathcal{P}_{\mathbf{x}} (u_k, \tau, {x_k, z_k, \dots}),
\end{align}
where $\mathbf{x}^{\text{aSLIP}}_{k+1} = [x_{k+1}, \dot{x}_{k+1}]^T$ is the horizontal state at pre-impact, $u_k$ is the previous step size, $\tau$ represents the leg length actuation over the step, and $x_k, z_k, \dots$ are the pre-impact states at the previous step. $\mathcal{P}_{\mathbf{x}}$ is guaranteed to exist since we  control the swing foot to periodically lift off and strike the ground in Eq. \eqref{eq:swingZpos}. The exact expression of $\mathcal{P}_{\mathbf{x}}$ cannot be obtained. We consider to approximate it via the S2S of the H-LIP in Eq. \eqref{eq:S2S}. Thus, Eq. \eqref{eq:SLIP_S2S} is rewritten as: 
\begin{align}
\label{eq:disturbedReturnMap}
\mathbf{x}^{\text{aSLIP}}_{k+1} &= A \mathbf{x}^{\text{aSLIP}}_{{k}}  +  B u_{k} + w,
\end{align}
where $w := \mathcal{P}_{\mathbf{x}}- A \mathbf{x}^{\text{aSLIP}}_{{k}}  -  B u_{k}$ can be viewed as the integral of the difference of the walking dynamics between the two models on the horizontal state over the step. Since each step happens in finite time, the integral is assumed to be bounded, i.e., $w \in W$ \rev{with $W$ being the set of $w$}. Let $u^{\text{H-LIP}}$ be the nominal step size to realize a desired walking behavior on the H-LIP, and $K$ is the feedback gain to make $A+BK$ stable. For the aSLIP, applying the \textit{H-LIP stepping}:
\begin{empheq}[box=\fbox]{align}
\label{eq:HLIPstepping}
u_k(\mathbf{x}^{\textrm{aSLIP}}_{k}) = u_k^{\textrm{H-LIP}} + K(\mathbf{x}^{\textrm{aSLIP}}_{k} - \mathbf{x}^{\textrm{H-LIP}}_{k} )
\end{empheq}
yields the \textit{error state} $\mathbf{e}_k := \mathbf{x}_k^{\text{aSLIP}} - \mathbf{x}_k^{\text{H-LIP}}$ to evolve on the error dynamics:
\begin{equation}
\label{closedLoopSystem}
\mathbf{e}_{k+1} = (A+BK) \mathbf{e}_{k}  + w,
\end{equation}
which has an \textit{error (disturbance) invariant set} $E$ by treating $w$ as the disturbance.  By definition,
$(A+BK) E  \oplus  W \in E$,
where $\oplus$ is the Minkowski sum. If $\mathbf{e}_0 \in E$, $\mathbf{e}_k \in E$, $ \forall k \in \mathbb{N}$. A small $W$ produces a small $E$. As a result, the H-LIP based stepping controller can approximately control the horizontal state of the aSLIP to exert the desired behavior with the tracking error being bounded by $E$. 
 Note that since the pre-impact state of the aSLIP $\mathbf{x}^{\textrm{aSLIP}}_{k}$ is not known as a priori, the desired step size is constantly calculated based on the current horizontal state $\mathbf{x}^{\textrm{aSLIP}}(t)$ in the SSP. 

\begin{algorithm}\caption{BBF-QP with H-LIP stepping for Walking}
 \begin{algorithmic}[1]
 \renewcommand{\algorithmicrequire}{\textbf{Initialization:}}
 \renewcommand{\algorithmicensure}{\textbf{Customization:}}
 \REQUIRE \textit{Terrain}: $z^{\text{terrain}}(x)$.
          \textit{Behavior}: $z_0 = 1$m, $T_\text{S} = 0.4$s, $T_\text{D} = 0.1$s. 
          \textit{Control}: $\alpha = 500$, $\gamma = 10$, $k= 10$, $c = 0.5$, $\Delta F = 20$. 
\WHILE {Simulation/Control loop}
\IF {SSP}
\STATE Desired step size $\leftarrow$ H-LIP stepping in Eq. \eqref{eq:HLIPstepping}
\STATE Desired swing foot position $\leftarrow$ Eq. \eqref{eq:swingXpos} \eqref{eq:swingZpos} \\
    \STATE $ \tau\leftarrow$ BBF-QP in Eq. \eqref{eq:BBF_QP_SSP}
\ELSE 
    \STATE $ \tau\leftarrow$ BBF-QP in Eq. \eqref{eq:BBF_QP_DSP}
\ENDIF
\ENDWHILE
 \end{algorithmic}
 \end{algorithm}
 
 \section{Results}
 %%% parameters
The control procedure \rev{with the chosen parameters is presented} in Algorithm 1. The stepping gain $K$ is chosen to be the deadbeat gain \cite{xiongInPrep} for all the walking for consistency, i.e., $(A +B K)^2 = 0$. The QP-based controller is solved at 1kHz. The aSLIP starts from an initial static configuration and walks to a desired pre-impact velocity $v^*$. The aSLIP parameters are chosen to match with the robot Cassie\footnote{$m = 33$kg, and $\text{K} = 8000$N/m, $\text{D} = 100$Ns/m, which are the nonlinear leg spring parameters of Cassie \cite{xiong2018bipedal} at $L \approx 1$m. The spring is chosen to be linear for generality.}. 
A video of the results can be seen in  \cite{supplementary}.% \href{https://youtu.be/fUZu6y-Gu4g}{\texttt{youtu.be/fUZu6y-Gu4g}}.}

 \begin{figure}[t]
      \centering
      \includegraphics[width = 1\columnwidth]{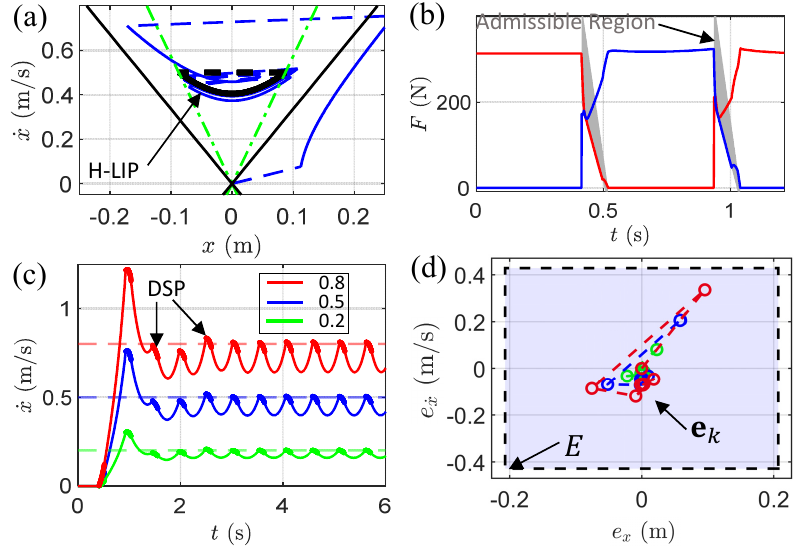} %done/flat2.pdf}%  inAir.jpg}
      \caption{Walking on flat ground: (a) the phase portrait of the horizontal state trajectory (blue) in SSP for walking with $v^* = 0.5$m/s, and the black is the corresponding orbit of the H-LIP \rev{(green lines are the orbital lines \cite{xiong2019IrosStepping}}, (b) the leg forces during walking with the \rev{gray} region being the admissible region in DSP; (c) the walking velocity and (d) the trajectories of the discrete error state $\mathbf{e}_k = [e_x, e_{\dot{x}}]^T$ of walking with different \rev{$v^* = 0.2, 0.5, 0.8$m/s (indicated by dashed lines in (c))}. }
      \label{fig:flat}
\end{figure}
 \begin{figure}[t]
      \centering
      \includegraphics[width = 0.95\columnwidth]{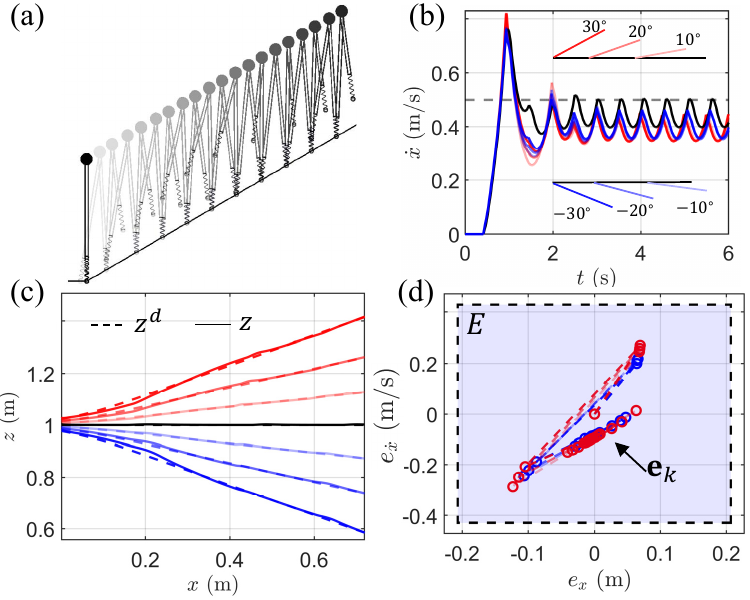} 
      \caption{Walking on slopes with $v^* = 0.5$m/s: (a) an illustration of the walking on an incline; (b) the velocity trajectories, (c) the vertical trajectories, and (d) the error state trajectories for walking on different slopes (the red plots indicate the inclines, the blue plots indicates the declines and the black plots represent the flat ground).}
      \label{fig:slope}
\end{figure}
 \begin{figure}[t]
      \centering
      \includegraphics[width = 1\columnwidth]{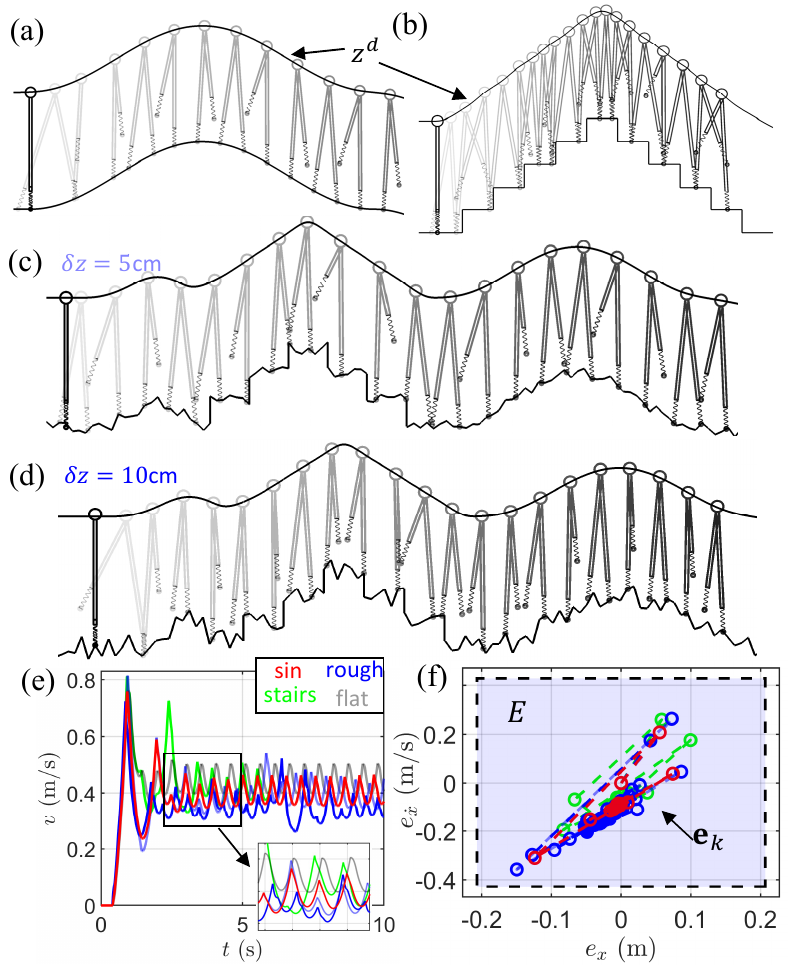}
      \caption{Walking with $v^* = 0.5$m/s on (a) sinusoidal terrain, (b) stairs, (c,d) rough terrains; (e) the velocity trajectories and (f) the error state trajectories.}
      \label{fig:rough}
\end{figure}
\subsection{Flat Ground and Slopes}

We first evaluate the approach for walking on flat ground and slopes, for which, the desired vertical trajectory is parallel to the terrain. Under this circumstance, the aSLIP best matches the original assumption of the H-LIP walking. %

Fig. \ref{fig:flat} shows the results of walking on flat ground. The aSLIP converges closely to the desired walking of the H-LIP. The leg forces behave as expected. Fig. \ref{fig:flat} (c) shows the trajectories of the horizontal velocity, which are not constant in the DSP and then contribute to the error $w$ in the S2S dynamics. We numerically calculate all $w$ for different walking simulations and inner approximate $W$ by a square. Since $(A+BK)^2 =0$, $E = W \oplus (A+BK)W$; we get an inner approximation of $E$ (shown in Fig. \ref{fig:flat} (d)), and all the error states $\mathbf{e}$ are inside $E$.

Fig. \ref{fig:slope} shows the walking on slopes up to $\pm30^{\circ}$: (b) the converged velocities are still close to the desired one,  (c) the vertical trajectories are controlled closely to the desired ones, and (d) the error states are inside $E$. \rev{The walking performance does not vary significantly on different slopes due to the trivial impact from the compliant spring.}

\subsection{Sin Waves, Stairs and Rough Terrains}
We then evaluate the walking on sin waves, stairs, and general rough terrains, as shown in Fig. \ref{fig:rough}. The desired vertical trajectories $z^d$ are not necessarily piece-wise linear and do not directly match the local slope assumption of the H-LIP, which presumably creates a larger $w$ and thus a looser tracking performance on the horizontal state. 

On sine waves, $z^d = z_0 + z^\text{terrain}$. For walking on stairs, we apply a moving averaging filter on $z^\text{terrain}$ to generate a smooth $z^d$. For walking on general rough terrains, we assume the terrain height is not exactly known. The terrain is generated with a combination of slopes, stairs, and sine waves plus a uniformly distributed noise with a maximum magnitude $\delta z$. We apply the moving averaging on the noise-free profile to get $z^d$. The noises are viewed as measurement errors from sensors on physical robots. We tested the cases with $\delta z = 0, 5, 10$cm. For even larger (unrealistic) noises, kinematic violation starts to happen, i.e., leg collides on the edges of the terrains.   

The results are shown in Fig. \ref{fig:rough}: the vertical trajectories are well-tracked, and the horizontal velocities (compared with the walking on flat ground) are tracked approximately. \rev{As expected, the walking on the rough terrains generated larger oscillations (larger $\delta z\rightarrow$ larger oscillations). This is in part because the noise creates variations on $T_\text{S}$, which contributes to $w$. Despite the velocity oscillations, the error states (blue dots in Fig. \ref{fig:rough}(f)) are all inside $E$.}

\section{Discussion}
The proposed approach is successfully realized on the aSLIP for controlling stable walking on various non-flat terrains. The performance of the walking and its robustness to the ground variation are encapsulated by the error invariant set. \rev{Despite the walking is presented in a plane, the approach can be readily applied to 3D walking of the aSLIP \cite{xiong2020ral} with two orthogonal planar stepping stabilizations via the H-LIP.}

The application of the H-LIP stepping relies on the condition that the aSLIP dynamics is close to that of the H-LIP. This is mainly ensured by three components of the control synthesis. The first is the direct control of the vertical state of the mass, so the vertical trajectory is not distant from that of the H-LIP. The second is the time-based component of the vertical swing foot trajectory, which makes sure that the duration of the SSP does not vary unless the terrain noise $\delta z$ is too large. The third is the control barrier function in DSP that guides one leg force to cross 0 at an appropriate timing, thus the DSP duration does not significantly vary. 

The assumption of constant velocity in DSP of the H-LIP is not problematic for walking with dominating SSP ($T_\text{S} > T_\text{D}$), as the integrated error over the DSP is small. The assumption can be improved for preciseness, e.g., by learning from the walking data. The horizontal state could be included in the BBF-QP as it is controllable when the leg is not vertical. However, it then becomes a balance or conflict of the control between the vertical and horizontal states.

\section{Conclusion and Future Work}
We present a \rev{highly efficient} control approach to enable actuated Spring Loaded Inverted Pendulum (aSLIP) to walking on rough terrains \rev{with large height variations}. The vertical state is controlled via Backstepping-Barrier function based quadratic programs (BBF-QPs); the horizontal state is stabilized via Hybrid-LIP based stepping. In the future, we will extend the approach to the aSLIP-like robots, e.g., Cassie and Digit, for walking over rough and challenging terrains.

\bibliographystyle{IEEEtran}
\bibliography{references}

\begin{thebibliography}{10}
\providecommand{\url}[1]{#1}
\csname url@samestyle\endcsname
\providecommand{\newblock}{\relax}
\providecommand{\bibinfo}[2]{#2}
\providecommand{\BIBentrySTDinterwordspacing}{\spaceskip=0pt\relax}
\providecommand{\BIBentryALTinterwordstretchfactor}{4}
\providecommand{\BIBentryALTinterwordspacing}{\spaceskip=\fontdimen2\font plus
\BIBentryALTinterwordstretchfactor\fontdimen3\font minus
  \fontdimen4\font\relax}
\providecommand{\BIBforeignlanguage}[2]{{%
\expandafter\ifx\csname l@#1\endcsname\relax
\typeout{** WARNING: IEEEtran.bst: No hyphenation pattern has been}%
\typeout{** loaded for the language `#1'. Using the pattern for}%
\typeout{** the default language instead.}%
\else
\language=\csname l@#1\endcsname
\fi
#2}}
\providecommand{\BIBdecl}{\relax}
\BIBdecl

\bibitem{raibert1986legged}
M.~H. Raibert, \emph{Legged robots that balance}.\hskip 1em plus 0.5em minus
  0.4em\relax MIT press, 1986.

\bibitem{full1999templates}
R.~J. Full and D.~E. Koditschek, ``Templates and anchors: neuromechanical
  hypotheses of legged locomotion on land,'' \emph{J. Exp. biol.}, vol. 202,
  no.~23, pp. 3325--3332, 1999.

\bibitem{geyer2006compliant}
H.~Geyer, A.~Seyfarth, and R.~Blickhan, ``Compliant leg behaviour explains
  basic dynamics of walking and running,'' \emph{Proc. Royal Soc. B}, vol. 273,
  no. 1603, pp. 2861--2867, 2006.

\bibitem{wuGeyer}
A.~{Wu} and H.~{Geyer}, ``Highly robust running of articulated bipeds in
  unobserved terrain,'' in \emph{2014 IEEE/RSJ Int. Conf. on Intell. Rob. and
  Sys. (IROS)}, pp. 2558--2565.

\bibitem{xiong2018bipedal}
X.~Xiong and A.~D. Ames, ``Bipedal hopping: Reduced-order model embedding via
  optimization-based control,'' in \emph{IEEE/RSJ Int. Conf. on Intell. Rob.
  and Sys. (IROS)}, 2018, pp. 3821--3828.

\bibitem{hutter2010slip}
M.~Hutter, C.~D. Remy, M.~A. H{\"o}pflinger, and R.~Siegwart, ``Slip running
  with an articulated robotic leg,'' in \emph{2010 IEEE/RSJ Int. Conf. on
  Intell. Rob. and Sys. (IROS)}, pp. 4934--4939.

\bibitem{rezazadeh2018robot}
S.~Rezazadeh, A.~Abate, R.~L. Hatton, and J.~W. Hurst, ``Robot leg design: A
  constructive framework,'' \emph{IEEE Access}, vol.~6, 2018.

\bibitem{haldane2017repetitive}
D.~W. Haldane, J.~K. Yim, and R.~S. Fearing, ``Repetitive extreme-acceleration
  (14-g) spatial jumping with salto-1p,'' in \emph{2017 IEEE/RSJ Int. Conf. on
  Intell. Rob. and Sys. (IROS)}, pp. 3345--3351.

\bibitem{GeyerChapter}
H.~Geyer and U.~Saranli, ``Gait based on the spring-loaded inverted pendulum,''
  \emph{In: Goswami A., Vadakkepat P. (eds) Humanoid Robotics: A Reference},
  2017.

\bibitem{Ahmadi2006ControlledPD}
M.~Ahmadi and M.~Buehler, ``Controlled passive dynamic running experiments with
  the arl-monopod ii,'' \emph{IEEE Trans. on Rob.}, vol.~22, pp. 974--986,
  2006.

\bibitem{ernst2010spring}
M.~Ernst, H.~Geyer, and R.~Blickhan, ``Spring-legged locomotion on uneven
  ground: a control approach to keep the running speed constant,'' in
  \emph{Mobile Robotics: Solutions and Challenges}.\hskip 1em plus 0.5em minus
  0.4em\relax World Scientific, 2010, pp. 639--644.

\bibitem{terry2015control}
P.~Terry and K.~Byl, ``Com control for underactuated 2d hopping robots with
  series-elastic actuation via higher order partial feedback linearization,''
  in \emph{2015 Proc IEEE Conf Decis Control}, pp. 7795--7801.

\bibitem{green2020planning}
K.~Green, R.~L. Hatton, and J.~Hurst, ``Planning for the unexpected: Explicitly
  optimizing motions for ground uncertainty in running,'' \emph{arXiv preprint
  arXiv:2001.10629}, 2020.

\bibitem{rezazadeh2015spring}
S.~Rezazadeh, C.~Hubicki, M.~Jones, A.~Peekema, J.~Van~Why, A.~Abate, and
  J.~Hurst, ``Spring-mass walking with atrias in 3d: Robust gait control
  spanning zero to 4.3 kph on a heavily underactuated bipedal robot,'' in
  \emph{ASME 2015 dynamic systems and control Conf.}

\bibitem{liu2016terrain}
Y.~Liu, P.~M. Wensing, J.~P. Schmiedeler, and D.~E. Orin, ``Terrain-blind
  humanoid walking based on a 3-d actuated dual-slip model,'' \emph{IEEE Rob.
  and Automation Letters}, vol.~1, no.~2, pp. 1073--1080, 2016.

\bibitem{piovan2015reachability}
G.~Piovan and K.~Byl, ``Reachability-based control for the active slip model,''
  \emph{I. J. Rob. Res.}, vol.~34, no.~3, pp. 270--287, 2015.

\bibitem{shemer2017flight}
N.~Shemer and A.~Degani, ``A flight-phase terrain following control strategy
  for stable and robust hopping of a one-legged robot under large terrain
  variations,'' \emph{Bioinspir Biomim}, vol.~12, no.~4, p. 046011, 2017.

\bibitem{dadashzadeh2014template}
B.~Dadashzadeh, H.~R. Vejdani, and J.~Hurst, ``From template to anchor: A novel
  control strategy for spring-mass running of bipedal robots,'' in \emph{2014
  IEEE/RSJ Int. Conf. on Intell. Rob. and Sys.(IROS)}, pp. 2566--2571.

\bibitem{xiong2020jumping}
X.~Xiong and A.~D. Ames, ``Sequential motion planning for bipedal somersault
  via flywheel slip and momentum transmission with task space control,'' in
  \emph{2020 IEEE/RSJ Int. Conf. on Intell. Rob. and Sys.}

\bibitem{xiong2019exo}
------, ``\href{https://arxiv.org/abs/1910.00687}{Motion Decoupling and
  Composition via Reduced Order Model Optimization for Dynamic Humanoid Walking
  with CLF-QP based Active Force Control},'' in \emph{In 2019 IEEE/RSJ Int.
  Conf. on Intell. Rob. and Sys. (IROS)}.

\bibitem{xiong2020ral}
------, ``Dynamic and versatile humanoid walking via embedding 3d actuated slip
  model with hybrid lip based stepping,'' \emph{IEEE Rob. and Automation
  Letters}, vol.~5, no.~4, pp. 6286--6293, 2020.

\bibitem{kokotovic}
P.~V. {Kokotovic}, ``The joy of feedback: nonlinear and adaptive,'' \emph{IEEE
  Control Systems Magazine}, vol.~12, no.~3, pp. 7--17, 1992.

\bibitem{ames2016control}
A.~D. Ames, X.~Xu, J.~W. Grizzle, and P.~Tabuada, ``Control barrier function
  based quadratic programs for safety critical systems,'' \emph{IEEE Trans. on
  Autom. Control}, vol.~62, no.~8, pp. 3861--3876, 2016.

\bibitem{xiong2019IrosStepping}
X.~Xiong and A.~D. Ames, ``Orbit characterization, stabilization and
  composition on 3d underactuated bipedal walking via hybrid passive linear
  inverted pendulum model,'' in \emph{2019 IEEE/RSJ Int. Conf. on Intell. Rob.
  and Sys. (IROS)}, pp. 4644--4651.

\bibitem{xiongInPrep}
X.~Xiong and A.~Ames, ``3d underactuated bipedal walking via h-lip based gait
  synthesis and stepping stabilization,'' \emph{arXiv preprint
  arXiv:2101.09588}, 2021.

\bibitem{dai2012optimizing}
H.~Dai and R.~Tedrake, ``Optimizing robust limit cycles for legged locomotion
  on unknown terrain,'' in \emph{2012 IEEE 51st IEEE Conf. on Decision and
  Control (CDC)}, pp. 1207--1213.

\bibitem{manchester2011stable}
I.~R. Manchester, U.~Mettin, F.~Iida, and R.~Tedrake, ``Stable dynamic walking
  over uneven terrain,'' \emph{I. J. Rob. Res.}, vol.~30, no.~3, 2011.

\bibitem{nguyen2018dynamic}
Q.~Nguyen, A.~Agrawal, W.~Martin, H.~Geyer, and K.~Sreenath, ``Dynamic bipedal
  locomotion over stochastic discrete terrain,'' \emph{Int. J. Robot. Res.},
  vol.~37, no. 13-14, pp. 1537--1553, 2018.

\bibitem{iida2010minimalistic}
F.~Iida and R.~Tedrake, ``Minimalistic control of biped walking in rough
  terrain,'' \emph{Autonomous Robots}, vol.~28, no.~3, pp. 355--368, 2010.

\bibitem{khalil1996noninear}
H.~K. Khalil, ``Nonlinear systems,'' \emph{Prentice-Hall, New Jersey}, 1996.

\bibitem{kajita2002}
S.~Kajita, F.~Kanehiro, K.~Kaneko, K.~Fujiwara, K.~Yokoi, and H.~Hirukawa, ``A
  realtime pattern generator for biped walking,'' in \emph{IEEE Int. Conf. on
  Rob. and Autom. (ICRA)}, vol.~1, 2002, pp. 31--37.

\bibitem{feng2016robust}
S.~Feng, X.~Xinjilefu, C.~G. Atkeson, and J.~Kim, ``Robust dynamic walking
  using online foot step optimization,'' in \emph{2016 IEEE/RSJ Int. Conf. on
  Intell. Rob. and Sys. (IROS)}, pp. 5373--5378.

\bibitem{kajita2001}
S.~{Kajita}, F.~{Kanehiro}, K.~{Kaneko}, K.~{Yokoi}, and H.~{Hirukawa}, ``The
  3d linear inverted pendulum mode: a simple modeling for a biped walking
  pattern generation,'' in \emph{2001 IEEE/RSJ Int. Conf. on Intell. Rob. and
  Sys. (IROS)}, pp. 239--246.

\bibitem{kajita1991study}
S.~Kajita and K.~Tani, ``Study of dynamic biped locomotion on rugged
  terrain-derivation and application of the linear inverted pendulum mode,'' in
  \emph{1991 IEEE Int. Conf. on Rob. and Autom (ICRA)}, pp. 1405--1406.

\bibitem{supplementary}
Simulation Results: [On-line] 2021. Available: \\
  \href{https://youtu.be/fUZu6y-Gu4g}{\texttt{https://youtu.be/fUZu6y-Gu4g}}.

\end{thebibliography}

\end{document}